\documentclass[twoside,11pt]{article}

\usepackage{jmlr2e}
\usepackage{lastpage}

\usepackage{cite}
\usepackage{adjustbox}

\renewcommand{\cite}[1]{\citep{#1}}
\usepackage{mathrsfs}
\usepackage{graphicx}

\usepackage{xcolor}

\newcommand{\metainfo}[1]{
	\noindent~\\
	\vspace{0.25cm}
	\fcolorbox{red}{green}{\parbox{0.97\textwidth}{#1\\}}
	\vspace{0.25cm}
}
\renewcommand{\metainfo}[1]{}
\newcommand{\crc}[1]{#1}
\newcommand{\book}[1]{#1}

\renewcommand{\book}[1]{}

\usepackage{amsmath}

\newcommand{\layer}[1]{L_{#1} }
\newcommand{\layeri}{\layer{i}}
\newcommand{\layerinput}{g}
\newcommand{\arch}{\mathsf{A}}
\newcommand{\aspace}{\mathcal{A}}

\usepackage{url}

\jmlrheading{20}{2019}{1-\pageref{LastPage}}{9/18; Revised 3/19}{3/19}{18-598}{Thomas Elsken, Jan Hendrik Metzen and Frank Hutter}

\firstpageno{1}

\begin{document}

	%
	\title{Neural Architecture Search: A Survey}
	%
	%
\author{\name Thomas Elsken \email thomas.elsken@de.bosch.com \\
       \addr Bosch Center for Artificial Intelligence\\
       71272 Renningen, Germany \\ and University of Freiburg  \\
       \AND
       \name Jan Hendrik Metzen \email JanHendrik.Metzen@de.bosch.com \\
        \addr Bosch Center for Artificial Intelligence\\
       71272 Renningen, Germany
       \AND
       \name Frank Hutter \email fh@cs.uni-freiburg.de \\
        \addr University of Freiburg\\
        79110 Freiburg, Germany}

\editor{Sebastian Nowozin}

	\maketitle              

	\begin{abstract}%
	    	Deep Learning has enabled remarkable progress over the last years on a variety of tasks, such as image recognition, speech recognition, and machine translation. One crucial aspect for this progress are novel neural architectures. Currently employed architectures have mostly been developed manually by human experts, which is a time-consuming and error-prone process. Because of this, there is growing interest in automated \emph{neural architecture search} methods. We provide an overview of existing work in this field of research and categorize them according to three dimensions: search space, search strategy, and performance estimation strategy.

	\end{abstract}


\setcounter{footnote}{0}

\section{Introduction} 
The success of deep learning in perceptual tasks is largely due to its automation of the feature engineering process: hierarchical feature extractors are learned in an end-to-end fashion from data rather than manually designed. This success has been accompanied, however, by a rising demand for \emph{architecture engineering}, where increasingly more complex neural architectures are designed manually. \emph{Neural Architecture Search} (NAS), the process of automating architecture engineering, is thus a logical next step in automating machine learning.
Already by now, NAS methods have outperformed manually designed architectures on some tasks such as image classification \citep{DBLP:journals/corr/ZophVSL17,real_regularized_2018}, object detection \citep{DBLP:journals/corr/ZophVSL17} or semantic segmentation \citep{NIPS2018_8087}.
NAS can be seen as subfield of AutoML~\cite{automl_book} and has significant overlap with hyperparameter optimization~\cite{feurer_hyperparameter_2018} and meta-learning~\cite{vanschoren_meta_2018}. We categorize methods for NAS according to three dimensions: search space, search strategy, and performance estimation strategy:

\begin{itemize}

\item[$\bullet$] \textbf{Search Space.} The search space defines which architectures can be represented in principle. Incorporating prior knowledge about typical properties of architectures well-suited for a task can reduce the size of the search space and simplify the search. However, this also introduces a human bias, which may prevent finding novel architectural building blocks that go beyond the current human knowledge. 

\item[$\bullet$] \textbf{Search Strategy.} The search strategy details how to explore the search space (which is often exponentially large or even unbounded). It encompasses the classical exploration-exploitation trade-off since, on the one hand, it is desirable to find well-performing architectures quickly, while on the other hand, premature convergence to a region of suboptimal architectures should be avoided. 

\item[$\bullet$] \textbf{Performance Estimation Strategy.} The objective of NAS is typically to find architectures that achieve high predictive performance on unseen data. \emph{Performance Estimation} refers to the process of estimating this performance: the simplest option is to perform a standard training and validation of the architecture on data, but this is unfortunately computationally expensive and limits the number of architectures that can be explored. Much recent research therefore focuses on developing methods that reduce the cost of these performance estimations.
\end{itemize}

\begin{figure}
		\centering
\includegraphics[width=1.0\textwidth]{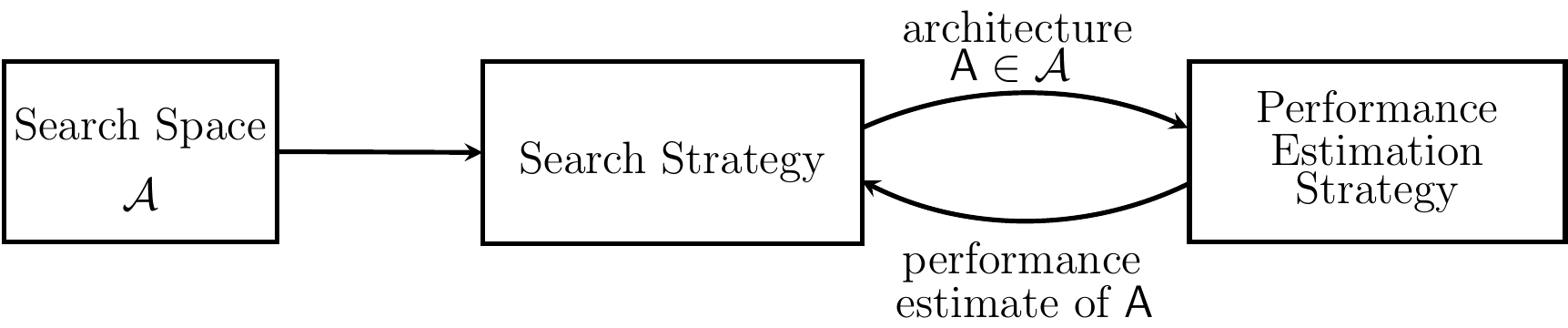}	
		\caption{Abstract illustration of Neural Architecture Search methods. A search strategy selects an architecture $\arch$ from a predefined search space $\aspace$. The architecture is passed to a performance estimation strategy, which returns the estimated performance of $\arch$ to the search strategy.}
		\label{fig:overview}
\end{figure}

We refer to Figure \ref{fig:overview} for an illustration. The 
article
is also structured according to these three dimensions: we start with discussing search spaces in Section \ref{Section:search_space}, cover search strategies in Section \ref{Section:search_heuristic}, and outline performance estimation methods in Section  \ref{Section:performance_estimation}. We conclude with an outlook on future directions in Section \ref{Section:outlook}.

\section{Search Space}\label{Section:search_space} 

The search space defines which neural architectures a NAS approach might discover in principle. We now discuss common search spaces from recent works.

\begin{figure}
		\centering
\includegraphics[scale=0.75]{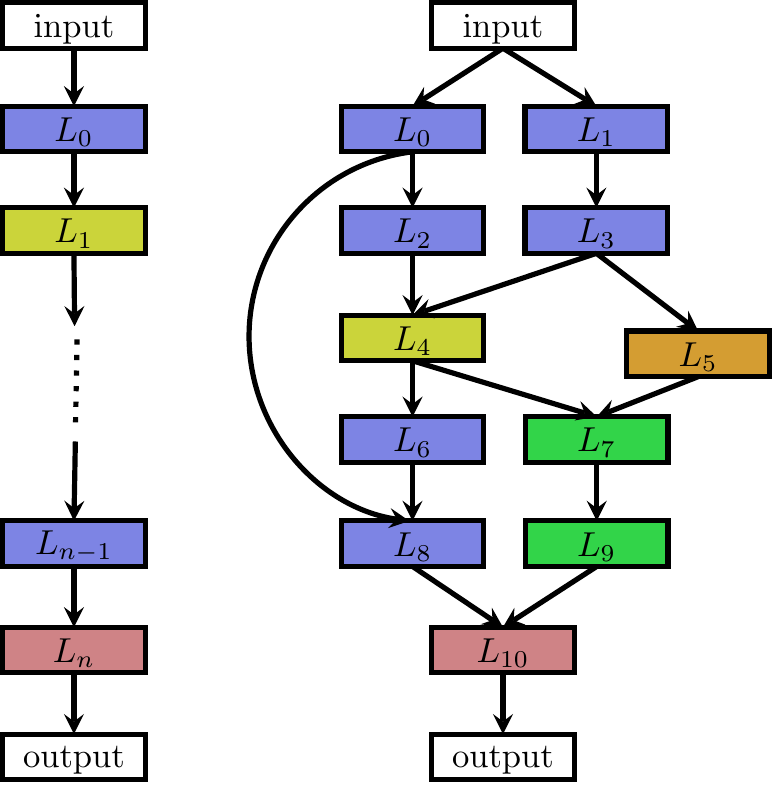}	
		\caption{An illustration of different architecture spaces. Each node in the graphs corresponds to a layer in a neural network, e.g., a convolutional or pooling layer. Different layer types are visualized by different colors. An edge from layer $\layer{i}$ to layer $\layer{j}$ denotes that $\layer{j}$ receives the output of $\layer{i}$ as input.  Left: an element of a chain-structured space. Right: an element of a more complex search space with additional layer types and multiple branches and skip connections. }
		\label{fig:chain_mb_space}
\end{figure}

A relatively simple search space is the space of \emph{chain-structured neural networks}, as illustrated in Figure \ref{fig:chain_mb_space} (left). A chain-structured neural network architecture $\arch$ can be written as a sequence of $n$ layers, where the i'th layer $\layeri$ receives its input from layer $i-1$ and its output serves as the input for layer $i+1$, i.e., $\arch = \layer{n} \circ \dots \layer{1} \circ \layer{0} $. The search space is then parametrized by: (i)~the (maximum) number of layers $n$ (possibly unbounded); (ii)~the type of operation every layer executes, e.g., pooling, convolution, or more advanced operations like depthwise separable convolutions \citep{DBLP:journals/corr/Chollet16a} or dilated convolutions \citep{YuKoltun2016}; and (iii)~hyperparameters associated with the operation, e.g., number of filters, kernel size and strides for a convolutional layer \citep{DBLP:journals/corr/BakerGNR16, DBLP:journals/corr/SuganumaSN17, cai2017reinforcement}, or simply number of units for fully-connected networks~\cite{mendoza_towards_2016}. Note that the parameters from (iii) are conditioned on (ii), hence the parametrization of the search space is not fixed-length but rather a conditional space. 

Recent work on NAS \citep{DBLP:journals/corr/abs-1708-05344,elsken_simple_2017,DBLP:journals/corr/ZophVSL17,elsken_multi_2018, real_regularized_2018,cai_path-level_2018} incorporates modern design elements known from hand-crafted architectures, such as skip connections, which allow to build complex, \emph{multi-branch networks}, as illustrated in Figure \ref{fig:chain_mb_space} (right). In this case the input of layer $i$ can be formally described as a function $\layerinput_i(\layer{i-1}^{out}, \dots , \layer{0}^{out})$ combining previous layer outputs. Employing such a function results in significantly more degrees of freedom. Special cases of these multi-branch architectures are (i) the chain-structured networks (by setting $\layerinput_i(\layer{i-1}^{out}, \dots , \layer{0}^{out}) = \layer{i-1}^{out}$), (ii) Residual Networks \citep{he_deep_2015}, where previous layer outputs are summed  ($\layerinput_i(\layer{i-1}^{out}, \dots , \layer{0}^{out}) = \layer{i-1}^{out}+\layer{j}^{out},j<i-1$) and (iii) DenseNets \citep{huang_densely_2016}, where previous layer outputs are concatenated ($\layerinput_i(\layer{i-1}^{out}, \dots , \layer{0}^{out}) = concat( \layer{i-1}^{out}, \dots , \layer{0}^{out})$).

\begin{figure}
		\centering
\includegraphics[scale=0.6]{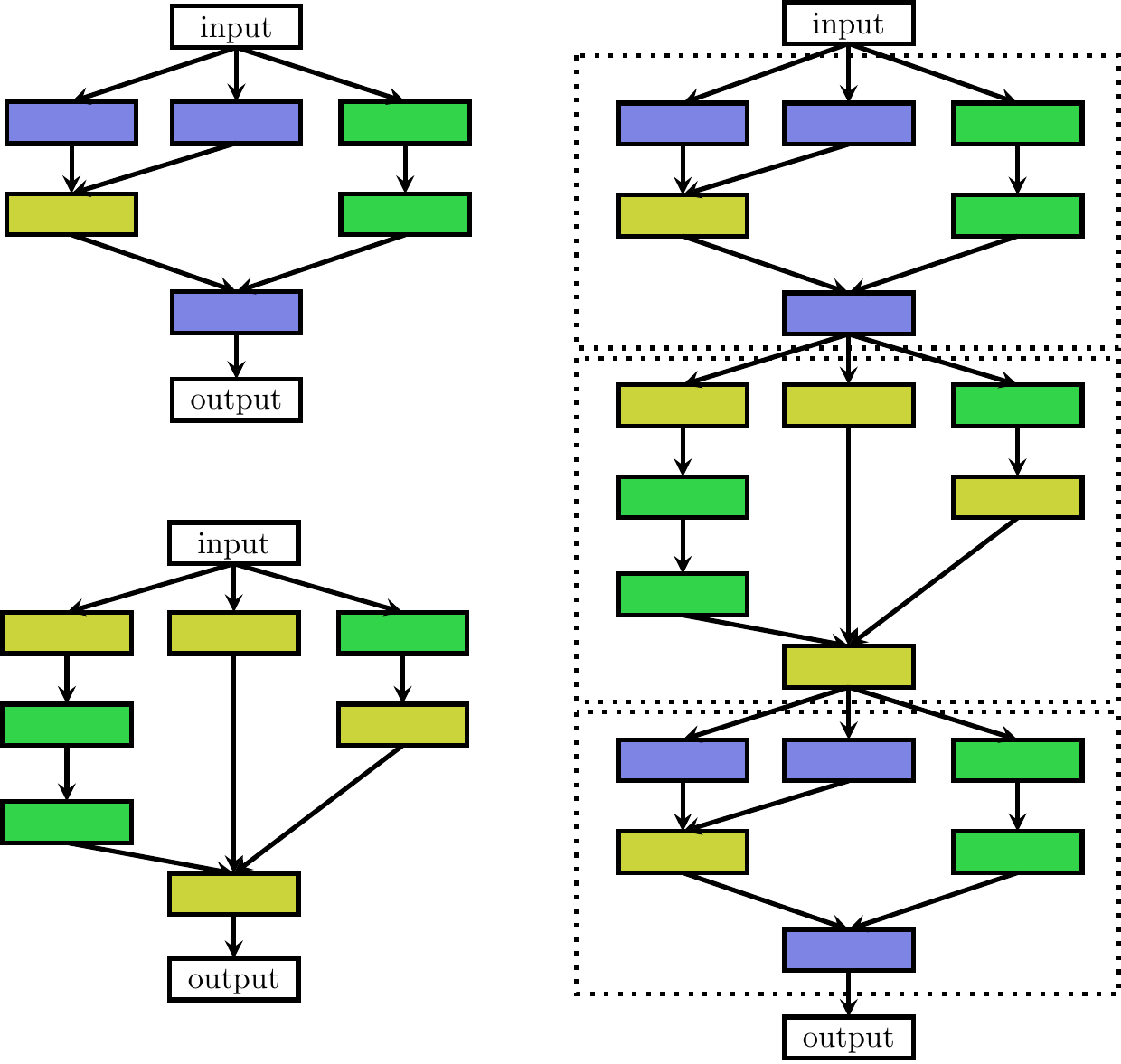}	
		\caption{Illustration of the cell search space. Left: Two different cells, e.g., a normal cell (top) and a reduction cell (bottom) \citep{DBLP:journals/corr/ZophVSL17}. Right: an architecture built by stacking the cells sequentially. Note that cells can also be combined in a more complex manner, such as in multi-branch spaces, by simply replacing layers with cells.}
		\label{fig:cell_space}
\end{figure}

Motivated by hand-crafted architectures consisting of repeated motifs  \citep{szegedy_rethinking_2015,he_deep_2015,huang_densely_2016}, \citet{DBLP:journals/corr/ZophVSL17} and \citet{zhong_practical_2017} propose to search for such motifs, dubbed \emph{cells} or \emph{blocks}, respectively, rather than for whole architectures. \citet{DBLP:journals/corr/ZophVSL17} optimize two different kind of cells: a \emph{normal cell} that preserves the dimensionality of the input and a \emph{reduction cell} which reduces the spatial dimension. The final architecture is then built by stacking these cells in a predefined manner, as illustrated in Figure \ref{fig:cell_space}. This search space has three major advantages compared to the ones discussed above: 
\begin{enumerate} 
\item The size of the search space is drastically reduced since cells usually consist of significantly less layers than whole architectures. For example, \citet{DBLP:journals/corr/ZophVSL17} estimate a seven-times speed-up compared to their previous work \citep{DBLP:journals/corr/ZophL16} while achieving better performance. 
\item  Architectures built from cells can more easily be transferred or adapted to other data sets by simply varying the number of cells and filters used within a model. Indeed, \citet{DBLP:journals/corr/ZophVSL17} transfer cells optimized on CIFAR-10 to ImageNet and achieve state-of-the-art performance.
\item 
Creating architectures by repeating building blocks has proven a useful design principle in general, such as repeating an LSTM block in RNNs or stacking a residual block.
\end{enumerate}

Consequently, this cell-based search space is also successfully employed by many recent works \citep{real_regularized_2018,liu_progressive_2017,pham_enas_2018,elsken_multi_2018,cai_path-level_2018, liu_darts,zhong2018blockqnn}. However, a new design-choice arises when using a cell-based search space, namely how to choose the \emph{macro-architecture}: how many cells shall be used and how should they be connected to build the actual model? For example, \citet{DBLP:journals/corr/ZophVSL17} build a sequential model from cells, in which each cell receives the outputs of the two preceding cells as input, while \citet{cai_path-level_2018} employ the high-level structure of well-known manually designed architectures, such as DenseNet \citep{huang_densely_2016}, and use their cells within these models. In principle, cells can be combined arbitrarily, e.g., within the multi-branch space described above, by simply replacing layers with cells. \crc{Ideally, both the macro-architecture and the micro-architecture (i.e., the structure of the cells) should be optimized jointly instead of solely optimizing the micro-architecture; otherwise, one may easily end up having to do manual macro-architecture engineering after finding a well-performing cell.
}
One step in the direction of optimizing macro-architectures is the hierarchical search space introduced by \citet{liu_hierarchical_2017}, which consists of several levels of motifs. The first level consists of the set of primitive operations, the second level of different motifs that connect primitive operations via a directed acyclic graph, the third level of motifs that encode how to connect second-level motifs, and so on. The cell-based search space can be seen as a special case of this hierarchical search space where the number of levels is three, the second level motifs correspond to the cells, and the third level is the hard-coded macro-architecture.

The choice of the search space largely determines the difficulty of the optimization problem: even for the case of the search space based on a single cell with fixed macro-architecture, the optimization problem remains (i) non-continuous and (ii) relatively high-dimensional (since more complex models tend to perform better, resulting in more design choices). 


We note that the architectures in many search spaces can be written as fixed-length vectors; e.g., the search space for each of the two cells by  \citet{DBLP:journals/corr/ZophVSL17} can be written as a 40-dimensional\footnote{Each of the 2 cells consists of 5 blocks. For each block, 2 inputs are chosen and for each of these inputs an operation to be applied to the input is chosen, resulting in a $2 \cdot 5 \cdot 2 \cdot 2 = 40$ dimensional representation. Note that originally there was another dimension for either summing or concatenating the operations within a block. However, this choice was discarded in the experiments of the paper and the outputs within a block were always summed.} search space with categorical dimensions, each of which chooses between a small number of different building blocks and inputs. 
Unbounded search spaces can be constrained to have a (potentially very large, but finite) number of layers, which again gives rise to fixed-size search spaces with (potentially many) conditional dimensions.

In the next section, we discuss Search Strategies that are well-suited for these kinds of search spaces.

\section{Search Strategy}\label{Section:search_heuristic} 

Many different search strategies can be used to explore the space of neural architectures, including random search, Bayesian optimization, evolutionary methods, reinforcement learning (RL), and gradient-based methods. Historically, evolutionary algorithms were already used by many researchers to evolve neural architectures (and often also their weights) decades ago~\citep[see, e.g.,][]{Angeline1994AnEA,stanley_evolving_2002, Floreano112676, Stanley:2009:HEE:1516090.1516093,pmlr-v37-jozefowicz15}. \citet{yao_evolving_1999} provides a literature review of work earlier than 2000.

Bayesian optimization celebrated several early successes in NAS since 2013, leading to state-of-the-art vision architectures~\cite{Bergstra2013MakingAS}, state-of-the-art performance for CIFAR-10 without data augmentation~\cite{DomSprHut15}, and the first automatically-tuned neural networks to win on competition data sets against human experts~\cite{mendoza_towards_2016}. 
NAS became a mainstream research topic in the machine learning community after \citet{DBLP:journals/corr/ZophL16} obtained competitive performance on the CIFAR-10 and Penn Treebank benchmarks with a search strategy based on reinforcement learning. While \citet{DBLP:journals/corr/ZophL16} used vast computational resources to achieve this result (800 GPUs for three to four weeks), after their work, a wide variety of methods have been published in quick succession to reduce the computational costs and achieve further improvements in performance.

To frame NAS as a \emph{reinforcement learning} (RL) problem \citep{DBLP:journals/corr/BakerGNR16, DBLP:journals/corr/ZophL16, zhong_practical_2017, DBLP:journals/corr/ZophVSL17}, the generation of a neural architecture can be considered to be the agent's action, with the action space identical to the search space. The agent's reward is based on an estimate of the performance of the trained architecture on unseen data (see Section \ref{Section:performance_estimation}). Different RL approaches differ in how they represent the agent's policy and how they optimize it: \citet{DBLP:journals/corr/ZophL16} use a recurrent neural network (RNN) policy to sequentially sample a string that in turn encodes the neural architecture. They initially trained this network with the REINFORCE policy gradient algorithm~\citep{Williams1992}, but in their follow-up work~\citep{DBLP:journals/corr/ZophVSL17} use Proximal Policy Optimization~\citep{DBLP:journals/corr/SchulmanWDRK17} instead. \citet{DBLP:journals/corr/BakerGNR16} use Q-learning to train a policy which sequentially chooses a layer's type and corresponding hyperparameters. 

An alternative view of these approaches is as sequential decision processes in which the policy samples actions to generate the architecture sequentially, the environment's ``state'' contains a summary of the actions sampled so far, and the (undiscounted) reward is obtained only after the final action. However, since no interaction with an environment occurs during this sequential process (no external state is observed, and there are no intermediate rewards), we find it more intuitive to interpret the architecture sampling process as the sequential generation of a single action; this simplifies the RL problem to a stateless multi-armed bandit problem.

A related approach was proposed by \citet{cai2017reinforcement}, who frame NAS as a sequential decision process: in their approach the state is the current (partially trained) architecture, the reward is an estimate of the architecture's performance, and the action corresponds to an application of function-preserving mutations, dubbed network morphisms \cite{DBLP:journals/corr/ChenGS15, DBLP:journals/corr/WeiWC17}, see also Section \ref{Section:performance_estimation}, followed by a training phase of the network. 
In order to deal with variable-length network architectures, they use a bi-directional LSTM to encode architectures into a fixed-length representation. Based on this encoded representation, actor networks decide on the sampled action. The combination of these two components constitute the policy, which is trained end-to-end with the REINFORCE policy gradient algorithm. We note that this approach will not visit the same state (architecture) twice.


An alternative to using RL are \emph{neuro-evolutionary} approaches that use evolutionary algorithms for optimizing the neural architecture. 
The first such approach for designing neural networks we are aware of dates back almost three decades: \citet{miller_designing_89} use genetic algorithms to propose architectures and use backpropagation to optimize their weights. 
Many neuro-evolutionary approaches since then \citep{Angeline1994AnEA,stanley_evolving_2002,Stanley:2009:HEE:1516090.1516093} use genetic algorithms to optimize both the neural architecture and its weights; however, when scaling to contemporary neural architectures with millions of weights for supervised learning tasks, SGD-based weight optimization methods currently outperform evolutionary ones.\footnote{Some recent work shows that evolving even millions of weights is competitive to gradient-based optimization when only high-variance estimates of the gradient are available, e.g., for reinforcement learning tasks \citep{Salimans2017_alternative,Such2017DeepNG,chrabaszcz-ijcai18a}. Nonetheless, for supervised learning tasks gradient-based optimization is by far the most common approach.} More recent neuro-evolutionary approaches \citep{DBLP:journals/corr/RealMSSSLK17, DBLP:journals/corr/SuganumaSN17,liu_hierarchical_2017,real_regularized_2018,miikkulainen_evolving_2017,xie_genetic_2017,elsken_multi_2018} therefore again use gradient-based methods for optimizing weights and solely use evolutionary algorithms for optimizing the neural architecture itself. 
Evolutionary algorithms evolve a population of models, i.e., a set of (possibly trained) networks; in every evolution step, at least one model from the population is sampled and serves as a parent to generate offsprings by applying mutations to it. In the context of NAS, mutations are local operations, such as adding or removing a layer, altering the hyperparameters of a layer, adding skip connections, as well as altering training hyperparameters. After training the offsprings, their fitness (e.g., performance on a validation set) is evaluated and they are added to the population.

Neuro-evolutionary methods differ in how they sample parents, update populations, and generate offsprings. For example, \citet{DBLP:journals/corr/RealMSSSLK17}, \citet{real_regularized_2018}, and \citet{liu_hierarchical_2017} use tournament selection \citep{Goldberg91acomparative} to sample parents, whereas \citet{elsken_multi_2018} sample parents from a multi-objective Pareto front using an inverse density. \citet{DBLP:journals/corr/RealMSSSLK17} remove the worst individual from a population, while \citet{real_regularized_2018} found it beneficial to remove the oldest individual (which decreases greediness), and \citet{liu_hierarchical_2017} do not remove individuals at all.
To generate offspring, most approaches initialize child networks randomly, while \citet{elsken_multi_2018} employ Lamarckian inheritance, i.e, knowledge (in the form of learned weights) is passed on from a parent network to its children by using network morphisms. \citet{DBLP:journals/corr/RealMSSSLK17} also let an offspring inherit all parameters of its parent that are not affected by the applied mutation; while this inheritance is not strictly function-preserving it might also speed up learning compared to a random initialization. Moreover, they also allow mutating the learning rate which can be seen as a way for optimizing the learning rate schedule during NAS.  We refer to \citet{stanley2019} for a recent in-depth review on neuro-evolutionary methods.

\citet{real_regularized_2018} conduct a case study comparing RL, evolution, and random search (RS), concluding that RL and evolution perform equally well in terms of final test accuracy, with evolution having better anytime performance and finding smaller models. Both approaches consistently perform better than RS in their experiments, but with a rather small margin: RS achieved test errors of approximately $4\%$ on CIFAR-10, while RL and evolution reached approximately $3.5\%$ (after ``model augmentation'' where depth and number of filters was increased; the difference on the non-augmented space actually used for the search was approx.\ 2\%). The difference was even smaller for \citet{liu_hierarchical_2017}, who reported a test error of $3.9\%$ on CIFAR-10 and a top-1 validation error of $21.0\%$ on ImageNet for RS, compared to $3.75\%$ and $20.3\%$ for their evolution-based method, respectively.

\emph{Bayesian Optimization} (BO, see, e.g., \cite{Shahriari_bo}) is one of the most popular methods for hyperparameter optimization\book{ (see also Chapter 1 of this book)}, but it has not been applied to NAS by many groups since typical BO toolboxes are based on Gaussian processes and focus on low-dimensional continuous optimization problems. \citet{SweDuvSnoHutOsb13} and \citet{kandasamy_neural_2018} derive kernel functions for architecture search spaces in order to use classic GP-based BO methods. 
In contrast, several works use tree-based models (in particular, tree Parzen estimators~\citep{Bergstra2011}, or random forests~\citep{hutter-lion11a}) to effectively search high-dimensional conditional spaces and achieve state-of-the-art performance on a wide range of problems, optimizing both neural architectures and their hyperparameters jointly~\cite{Bergstra2013MakingAS,DomSprHut15,mendoza_towards_2016,Zela_AutoML2018}.
While a full comparison is lacking, there is preliminary evidence that these approaches can also outperform evolutionary algorithms~\cite{klein_rml2018}. 

\citet{Negrinho_deeparch} and \citet{wistuba_finding_2017} exploit the tree-structure of their search space and use \emph{Monte Carlo Tree Search}. \citet{elsken_simple_2017} propose a simple yet well performing \emph{hill climbing} algorithm that discovers high-quality architectures by greedily moving in the direction of better performing architectures without requiring more sophisticated exploration mechanisms. 

While the methods above employ a discrete search space, \citet{liu_darts} propose a \emph{continuous relaxation} to enable direct gradient-based optimization:
instead of fixing a single operation $o_i$ (e.g., convolution or pooling) to be executed at a specific layer, the authors compute a convex combination from a set of operations $\{o_1, \dots, o_m \}$. More specifically, given a layer input $x$, the layer output $y$ is computed as $y  = \sum_{i=1}^{m} \alpha_i o_i(x), \alpha_i \ge 0, \sum_{i=1}^m \alpha_i = 1 $, where the convex coefficients $\alpha_i$ effectively parametrize the network architecture. \citet{liu_darts} then optimize both the network weights and the network architecture by alternating gradient descent steps on training data for weights and on validation data for architectural parameters such as $\alpha$. Eventually, a discrete architecture is obtained by choosing the operation $i^*$ with $ i^* =  {\arg\max}_i \,  \alpha_i$ for every layer. Instead of optimizing a weighting $\alpha$ of possible operations, \citet{xie2018snas,cai2018proxylessnas} propose to optimize a parametrized distribution over the possible operations. \citet{shin2018differentiable} and \citet{ahmed_connect} also employ gradient-based optimization of neural architectures, however focusing on optimizing layer hyperparameters or connectivity patterns, respectively. 

\section{Performance Estimation Strategy}\label{Section:performance_estimation} 

The search strategies discussed in Section \ref{Section:search_heuristic} aim at finding a neural architecture $\arch$ that maximizes some performance measure, such as accuracy on unseen data. To guide their search process, these strategies need to estimate the performance of a given architecture $\arch$ they consider. The simplest way of doing this is to train $\arch$ on training data and evaluate its performance on validation data. 
However, training each architecture to be evaluated from scratch frequently yields computational demands in the order of thousands of GPU days for NAS  \citep{DBLP:journals/corr/ZophL16,DBLP:journals/corr/RealMSSSLK17,DBLP:journals/corr/ZophVSL17,real_regularized_2018}. \crc{This naturally leads to developing methods for speeding up performance estimation, which we will now discuss. We refer to Table \ref{tbl:speedup} for an overview of existing methods.}

\begin{table}[]\label{tbl:speedup} 
\begin{tabular}{|l |l|l|}
\hline
\multicolumn{1}{|c|}{\textbf{Speed-up method}}          & \textbf{How are speed-ups achieved?}                                                                                                                                                                   & \textbf{References}               \\ \hline
\textbf{\begin{tabular}[c]{@{}l@{}}Lower fidelity\\ estimates\end{tabular}}       & \begin{tabular}[c]{@{}l@{}}Training time reduced by\\training for fewer epochs, on\\subset of data, downscaled\\models, downscaled data, ...\end{tabular}                             & \begin{tabular}[c]{@{}l@{}}
 \citet{hyperband}, \\ \citet{DBLP:journals/corr/ZophVSL17}, \\ 
 \citet{Zela_AutoML2018}, \\ \citet{pmlr-v80-falkner18a}, \\ \citet{real_regularized_2018},\\ \citet{runge2019learning} \end{tabular} \\ \hline
\textbf{\begin{tabular}[c]{@{}l@{}}Learning Curve\\ Extrapolation\end{tabular}}    & \begin{tabular}[c]{@{}l@{}}Training time reduced as \\ performance can be extrapolated \\ after just a few epochs of training.\end{tabular}                                                                 & \begin{tabular}[c]{@{}l@{}} 
\citet{swersky_freezethaw_2014}, \\ \citet{DomSprHut15},\\ \citet{klein-iclr17}, \\ \citet{baker_accelerating_2017}
\end{tabular}         \\ \hline
\textbf{\begin{tabular}[c]{@{}l@{}}Weight Inheritance/ \\ Network Morphisms\end{tabular}}             & \begin{tabular}[c]{@{}l@{}}Instead of training models from \\scratch, they are warm-started by \\inheriting weights of, e.g.,  a\\parent model.\end{tabular}                                               & \begin{tabular}[c]{@{}l@{}}
\citet{DBLP:journals/corr/RealMSSSLK17},\\ \citet{elsken_simple_2017}, \\ \citet{elsken_multi_2018}, \\ \citet{cai2017reinforcement, cai_path-level_2018} \\

\end{tabular}       \\ \hline
\textbf{\begin{tabular}[c]{@{}l@{}}One-Shot Models/\\ Weight Sharing\end{tabular}} & \begin{tabular}[c]{@{}l@{}}Only the one-shot model needs \\to be trained; its weights are \\ then shared across different \\architectures that are just \\subgraphs of the one-shot model.\end{tabular} & \begin{tabular}[c]{@{}l@{}}
\small{\citet{DBLP:journals/corr/SaxenaV16}}, \\\citet{pham_enas_2018}, \\ \citet{bender_icml:2018},\\ \citet{liu_darts}, \\ \citet{cai2018proxylessnas}, \\ \citet{xie2018snas}

\end{tabular}       \\ \hline
\end{tabular}
\caption{Overview of different methods for speeding up performance estimation in NAS.}
\vspace*{-0.5cm}
\end{table}

Performance can be estimated based on \emph{lower fidelities} of the actual performance after full training (also denoted as proxy metrics). Such lower fidelities include shorter training times \cite{DBLP:journals/corr/ZophVSL17,Zela_AutoML2018}, training on a subset of the data \cite{pmlr-v54-klein17a}, on lower-resolution images~\cite{Chrabaszcz_arXiv17}, or with less filters per layer and less cells \cite{DBLP:journals/corr/ZophVSL17,real_regularized_2018}. While these low-fidelity approximations reduce the computational cost, they also introduce bias in the estimate as performance will typically be underestimated. This may not be problematic as long as the search strategy only relies on ranking different architectures and the relative ranking remains stable. However, recent results indicate that this relative ranking can change dramatically when the difference between the cheap approximations and the ``full'' evaluation is too big~\cite{Zela_AutoML2018}, arguing for a gradual increase in fidelities~\cite{hyperband,pmlr-v80-falkner18a}.

Another possible way of estimating an architecture's performance builds upon learning curve extrapolation~\cite{swersky_freezethaw_2014,DomSprHut15,klein-iclr17,baker_accelerating_2017,rawal_nodes_2018}.
\citet{DomSprHut15} propose to extrapolate initial learning curves and terminate those predicted to perform poorly to speed up the architecture search process. \citet{swersky_freezethaw_2014}, \citet{klein-iclr17}, \citet{baker_accelerating_2017}, and \citet{rawal_nodes_2018} also consider architectural hyperparameters for predicting which partial learning curves are most promising. 
Training a surrogate model for predicting the performance of novel architectures is also proposed by \citet{liu_progressive_2017}, who do not employ learning curve extrapolation but support predicting performance based on architectural/cell properties and extrapolate to architectures/cells with larger size than seen during training. The main challenge for predicting the performances of neural architectures is that, in order to speed up the search process, good predictions in a relatively large search space need to be made based on relatively few evaluations. 

Another approach to speed up performance estimation is to initialize the weights of novel architectures based on weights of other architectures that have been trained before. One way of achieving this, dubbed \emph{network morphisms} \citep{DBLP:journals/corr/WeiWRC16}, allows modifying an architecture while leaving the function represented by the network unchanged, 
resulting in methods \crc{that only require a few GPU days} \cite{elsken_simple_2017,cai2017reinforcement, cai_path-level_2018,jin2018efficient}. This allows increasing the capacity of networks successively and retaining high performance without requiring training from scratch. Continuing training for a few epochs can also make use of the additional capacity introduced by network morphisms. An advantage of these approaches is that they allow search spaces without an inherent upper bound on the architecture's size \cite{elsken_simple_2017}; on the other hand, strict network morphisms can only make architectures larger and may thus lead to overly complex architectures. This can be attenuated by employing approximate network morphisms that allow shrinking architectures \cite{elsken_multi_2018}. 

\begin{figure}	\vspace*{-1.0cm}
		\centering
\includegraphics[width=.80\textwidth]{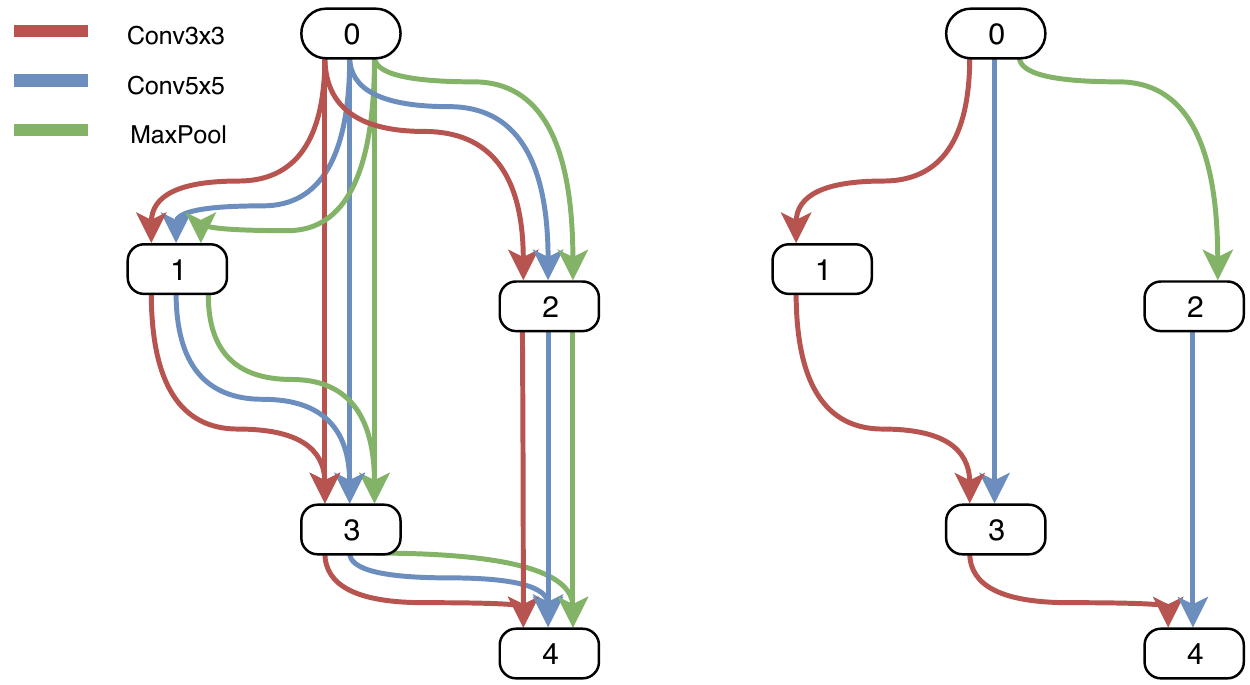}	
		\caption{ \crc{Illustration of one-shot architecture search. Simple network with an input node (denoted as 0), three hidden nodes (denoted as 1,2,3) and one output node (denoted as 4). Instead of applying a single operation (such as a 3x3 convolution) to a node, the one-shot model (left) contains several candidate operations for every node, namely 3x3 convolution (red edges), 5x5 convolution (blue edges) and MaxPooling (green edges) in the above illustration. Once the one-shot model is trained, its weights are shared across different architectures, which are simply subgraphs of the one-shot model (right).	Figure inspired by \citet{liu_darts}.}}
		\label{fig:one-shot}
\end{figure}

\emph{One-Shot Architecture Search} (see Figure \ref{fig:one-shot}) treats all architectures as different subgraphs of a supergraph (the one-shot model) and shares weights between architectures that have edges of this supergraph in common \cite{DBLP:journals/corr/SaxenaV16, DBLP:journals/corr/abs-1708-05344, pham_enas_2018,liu_darts,bender_icml:2018,cai2018proxylessnas,xie2018snas}. Only the weights of a single one-shot model need to be trained (in one of various ways), and architectures (which are just subgraphs of the one-shot model) can then be evaluated without any separate training by inheriting trained weights from the one-shot model. This greatly speeds up performance estimation of architectures, since no training is required (only evaluating performance on validation data)\crc{, again resulting in methods that only require a few GPU days}. The one-shot model typically incurs a large bias as it underestimates the actual performance of the best architectures severely; nevertheless, it allows ranking architectures, which would be sufficient if the estimated performance correlates strongly with the actual performance. However, it is currently not clear if this is actually the case \cite{bender_icml:2018, sciuto_evalnas}.

Different one-shot NAS methods differ in how the one-shot model is trained: ENAS \citep{ pham_enas_2018} learns a RNN controller that samples architectures from the search space and trains the one-shot model based on approximate gradients obtained through REINFORCE. DARTS \cite{ liu_darts} optimizes all weights of the one-shot model jointly with a continuous relaxation of the search space, obtained by placing a mixture of candidate operations on each edge of the one-shot model. \crc{Instead of optimizing real-valued weights on the operations as in DARTS, SNAS \cite{xie2018snas} optimizes a distribution over the candidate operations. The authors employ the concrete distribution~\citep{maddison_concrete,jang_gumbel} and reparametrization~\citep{kingma_repara} to relax the discrete distribution and make it differentiable, enabling optimization via gradient descent. To overcome the necessity of keeping the entire one-shot model in the GPU memory, ProxylessNAS \cite{cai2018proxylessnas} ``binarizes'' the architectural weights, masking out all but one edge per operation. The probabilities for an edge being either masked out or not are then learned by sampling a few binarized architectures and using BinaryConnect~\citep{Courbariaux_bc} to update the corresponding probabilities.} \citet{ bender_icml:2018} only train the one-shot model once and show that this is sufficient when deactivating parts of this model stochastically during training using path dropout. 

While the previously mentioned approaches optimize a distribution over architectures during training, the approach of \citet{ bender_icml:2018} can be seen as using a fixed distribution. The high performance obtainable by the latter indicates that the combination of weight sharing and a fixed (carefully chosen) distribution might (perhaps surprisingly) be the only required ingredients for one-shot NAS.
Related to these approaches is meta-learning of hypernetworks that generate weights for novel architectures and thus requires only training the hypernetwork but not the architectures themselves \cite{DBLP:journals/corr/abs-1708-05344,zhang2018graph}. The main difference here is that weights are not strictly shared but generated by the shared hypernetwork (conditional on the sampled architecture).   

A general limitation of one-shot NAS is that the supergraph defined a priori restricts the search space to its subgraphs. Moreover, approaches which require that the entire supergraph resides in GPU memory during architecture search will be restricted to relatively small supergraphs and search spaces accordingly, and are thus typically used in combination with cell-based search spaces. \crc{While approaches based on weight sharing have substantially reduced the computational resources required for NAS (from thousands to a few GPU days), it is currently not well understood which biases they introduce into the search if the sampling distribution of architectures is optimized along with the one-shot model instead of fixing it~\citep{bender_icml:2018}. For instance, an initial bias in exploring certain parts of the search space more than others might lead to the weights of the one-shot model being better adapted for these architectures, which in turn would reinforce the bias of the search to these parts of the search space. This might result in premature convergence of NAS or little correlation between the one-shot and true performance of an architecture~\citep{sciuto_evalnas}}. In general, a more systematic analysis of biases introduced by different performance estimators would be a desirable direction for future work.

\section{Future Directions}\label{Section:outlook} 
In this section, we discuss several current and future directions for research on NAS. Most existing work has focused on NAS for image classification. On the one hand, this provides a challenging benchmark since a lot of manual engineering has been devoted to finding architectures that perform well in this domain and are not easily outperformed by NAS. On the other hand, it is relatively easy to define a well-suited search space by exploiting knowledge from manual engineering. This in turn makes it unlikely that NAS will find architectures that substantially outperform existing ones considerably since the found architectures cannot differ fundamentally. We thus consider it important to go beyond image classification problems by applying NAS to less explored domains. 
Notable first steps in this direction have been taken in image restoration~\cite{pmlr-v80-suganuma18a}, semantic segmentation~\citep{NIPS2018_8087, Nekrasov_semseg, liu_autodeeplab}, transfer learning~\citep{NIPS2018_8056}, machine translation \cite{so_transfomrer}, 
reinforcement learning~\cite{runge2019learning}\footnote{Many authors have optimized some architectural choices of deep reinforcement learning algorithms before. \citet{runge2019learning} used the largest and most versatile space of policy network architectures so far (e.g., including both convolutional and recurrent building blocks), but a full study of NAS for RL is yet missing.}, as well as optimizing recurrent neural networks \cite{greff_2015, pmlr-v37-jozefowicz15, DBLP:journals/corr/ZophL16,rawal_nodes_2018}, e.g., for language or music modeling. Further promising application areas for NAS would be generative adversarial networks or sensor fusion.

An additional promising direction is to develop NAS methods for multi-task problems \cite{liang_evolutionary_2018, meyerson_pseudo-task_2018} and for multi-objective problems \cite{elsken_multi_2018, dong_dpp_net_2018,zhou_renas}, in which measures of resource efficiency are used as objectives along with the predictive performance on unseen data. 
We highlight that multi-objective NAS is closely related to network compression~\citep{han_compression,cheng_compression}: both aim at finding well-performing but efficient architectures. Hence, some compression methods can also be seen as NAS methods~\citep{NIPS2015_5784, Liu2017LearningEC,Gordon_2018_CVPR,liu2018rethinking,cao2018learnable} and vice versa~\citep{DBLP:journals/corr/SaxenaV16,liu_darts,xie2018snas}. 

Likewise, it would be interesting to extend RL/bandit approaches, such as those discussed in Section \ref{Section:search_heuristic}, to learn policies that are conditioned on a state that encodes task properties/resource requirements (i.e., turning the setting into a contextual bandit). A similar direction was followed by \citet{ramachandran_dynamic_2018} in extending one-shot NAS to generate different architectures depending on the task or instance on-the-fly. Moreover, applying NAS to searching for architectures that are more robust to adversarial examples \cite{cubuk_intriguing_2017} is an intriguing recent direction.

Related to this is research on defining more general and flexible search spaces. For instance, while the cell-based search space provides high transferability between different image classification tasks, it is largely based on human experience on image classification and does not generalize easily to other domains where the hard-coded hierarchical structure (repeating the same cells several times in a chain-like structure) does not apply (e.g., semantic segmentation or object detection). A search space which allows representing and identifying more general hierarchical structure would thus make NAS more broadly applicable, see \citet{liu_hierarchical_2017,liu_autodeeplab} for first work in this direction. Moreover, common search spaces are also based on predefined building blocks, such as different kinds of convolutions and pooling, but do not allow identifying novel building blocks on this level; going beyond this limitation might substantially increase the power of NAS.

The comparison of different methods for NAS \crc{and even the reproducibility of published results\footnote{We refer to  \citet{li_repro_nas} for a detailed discussion on reproducibility for NAS. }} is complicated by the fact that measurements of an architecture's performance depend on many factors other than the architecture itself. While most authors report results on the CIFAR-10 data set, experiments often differ with regard to search space, computational budget, data augmentation, training procedures, regularization, and other factors. For example, for CIFAR-10, performance substantially improves when using a cosine annealing learning rate schedule \citep{DBLP:journals/corr/LoshchilovH16a}, data augmentation by CutOut \citep{cutout}, by MixUp \citep{mixup} or by a combination of factors \citep{cubuk_autoaugment:_2018}, and regularization by Shake-Shake regularization \citep{DBLP:journals/corr/Gastaldi17} or ScheduledDropPath \citep{DBLP:journals/corr/ZophVSL17}.
It is therefore conceivable that improvements in these ingredients have a larger impact on reported performance numbers than the better architectures found by NAS. We thus consider the definition of common benchmarks to be crucial for a fair comparison of different NAS methods. \crc{A first step in this direction is the benchmark proposed by \citet{aaron_nasbench}, where a search space consisting of approximately 423,000 unique convolutional architectures is considered. Each element of this space was pre-trained and evaluated multiple times, resulting in a data set containing training, validation and test accuracies as well as training times and model sizes for different training budgets for multiple runs. Different search strategies can hence be compared with low computational resources on this benchmark by simply querying the pre-computed data set. In a smaller previous study, ~\citet{klein_rml2018} pre-evaluated the joint space of neural architectures and hyperparameters.}
It would also be interesting to evaluate NAS methods not in isolation but as part of a full open-source AutoML system, where also hyperparameters \cite{mendoza_towards_2016, DBLP:journals/corr/RealMSSSLK17, Zela_AutoML2018}, and data augmentation pipeline \cite{cubuk_autoaugment:_2018} are optimized along with NAS.

While NAS has achieved impressive performance, so far it provides little insights into why specific architectures work well and how similar the architectures derived in independent runs would be. Identifying common motifs, providing an understanding why those motifs are important for high performance, and investigating if these motifs generalize over different problems would be desirable. 

\acks{We would like to thank Arber Zela, Esteban Real, Gabriel Bender, Kenneth Stanley, Thomas Pfeil and the anonymous reviewers for feedback on this survey.
This work has partly been supported by the European Research Council (ERC) under the European Union’s Horizon 2020 research and innovation programme under grant no.\ 716721.}

\bibliography{as_chapter}
\bibliographystyle{spbasic}

\metainfo{2 pages}
\end{document}